%% file: main.tex
\documentclass[11pt,a4paper]{lumia}

\usepackage{natbib}
\setcitestyle{authoryear,round,citesep={;},aysep={,},yysep={;}}
\input{math_commands.tex}

\usepackage{multirow}
\usepackage{ragged2e}
\usepackage{placeins}
\usepackage{etoolbox}
\input{paper_macros.tex}

\patchcmd{\maketitle}
  {\begin{adjustwidth}{0pt}{24pt}}
  {\begin{adjustwidth}{0pt}{0pt}}
  {}{\PackageError{lumia-layout}{Could not adjust the title block width}{}}

\setheadertext{LUMIA Lab}
\setheadertitle{Deep Fusion Memory}
\title{Deep Fusion Memory: A Chunk-Level\\Parametric Memory for Large Language Models}

\author{%
  Rubin Wei$^{1,2}$, Siyuan Sun$^{1}$, Hao Sun$^{1,2}$, Yuliang Liu$^{1,2}$, Xiang Shi$^{1}$, Jiarui Wang$^{1,2}$, Junming Zhang$^{1,2}$,\\
  \Authfont Zhouhan Lin$^{1,2\ddagger}$\\
  {\Affilfont $^1$ LUMIA Lab, School of Artificial Intelligence, Shanghai Jiao Tong University}\\
  {\Affilfont $^2$ Shanghai Artificial Intelligence Laboratory}
}
\correspondingemail{\emailicon\ weirubinn@gmail.com, lin.zhouhan@gmail.com \quad $^\ddagger$ Corresponding Author.}
\githublink{}
\huggingfacelink{}
\hypersetup{pdftitle={Deep Fusion Memory: A Chunk-Level Parametric Memory for Large Language Models},pdfauthor={Rubin Wei, Siyuan Sun, Hao Sun, Yuliang Liu, Xiang Shi, Jiarui Wang, Junming Zhang, Zhouhan Lin}}

\begin{document}

\begin{abstract}
\input{sections/00_abstract}
\end{abstract}
\maketitle

\input{sections/01_introduction}
\input{sections/02_related_work}
\input{sections/03_method}
\input{sections/04_experiments}
\input{sections/05_analysis_and_ablations}
\input{sections/05_conclusion}
\input{sections/06_statements}

\bibliography{references}

\clearpage
\appendix
\input{sections/08_appendix}

\end{document}

%% file: math_commands.tex
\usepackage{amsmath,amsfonts,bm}

\def\eqref#1{equation~\ref{#1}}

\def\1{\bm{1}}

\DeclareMathAlphabet{\mathsfit}{\encodingdefault}{\sfdefault}{m}{sl}
\SetMathAlphabet{\mathsfit}{bold}{\encodingdefault}{\sfdefault}{bx}{n}



%% file: paper_macros.tex
\newcommand{\method}{Deep Fusion Memory}

\definecolor{dfmhighlight}{cmyk}{0.124,0,0.024,0}
\definecolor{dfmgroup}{gray}{0.96}
\definecolor{dfmgain}{rgb}{0,0.6,0}
\definecolor{dfmloss}{rgb}{1,0.2,0.2}
\newcommand{\scoredecrease}[2]{#1\rlap{$_{\textcolor{dfmloss}{\text{#2}}}$}}
\definecolor{evidencecolor}{HTML}{087F8C}
\definecolor{memorybg}{HTML}{F2F8FA}
\definecolor{targetbg}{HTML}{F7FAF5}

\newcolumntype{Y}{>{\RaggedRight\arraybackslash\hspace{0pt}}X}

%% file: sections/00_abstract.tex
Large language models need effective ways to acquire and use external knowledge beyond what they retain from pretraining. Continued pretraining risks catastrophic forgetting, while retrieval at inference time can add substantial overhead. External parametric memory avoids online retrieval, but constructing its training data requires extensive offline search, and output-level interpolation provides only shallow fusion. We introduce \method{} (DFM), a parametric memory architecture that reduces this construction workload and incorporates learned memory into intermediate hidden states. A causal memory predictor learns to generate memory from the preceding context, using retrieved chunk embeddings from the corpus as supervision. The predicted memory is shared across tokens within each chunk and fused into their hidden states through gated cross-attention. Memory prediction and fusion are trained end to end with the backbone frozen, and inference requires no explicit retrieval. Compared with existing parametric memory approaches, which often construct token-level memory targets, DFM requires $32\times$ fewer retrieval queries and $2048\times$ fewer retrieved items to construct its training data. Across six QA tasks, DFM performs better over the strongest baseline by 12.71\% on Mistral-7B-v0.3 and 11.19\% on Qwen3-8B-Base. Beyond QA, domain adaptation yields average gains of 6.51 and 4.38 points over Qwen3-4B-Base and Qwen3-8B-Base, respectively.
The code will be released soon.

%% file: sections/01_introduction.tex
\section{Introduction}
\label{sec:introduction}

Large language models (LLMs)~\citep{yang2025qwen3,deepseekai2025deepseekv32,kimiteam2026kimik2,bai2026interns2} have advanced rapidly in language understanding, generation, and reasoning. Their growing capabilities support knowledge-intensive applications, but reliable answers require facts that can be missing, poorly retained, or outdated in pretrained parameters~\citep{lewis2020rag}. Knowledge injection addresses this gap by incorporating information from an external corpus. A central challenge is to incorporate this knowledge effectively while preserving existing capabilities and keeping knowledge acquisition and inference costs manageable.

Existing approaches face different trade-offs in addressing this challenge. Continued pretraining (CPT)~\citep{gururangan2020dont} learns from an additional corpus through language modeling, while LoRA~\citep{hu2022lora} offers a parameter-efficient way to adapt the model. These approaches keep inference self-contained, but adaptation can cause catastrophic forgetting of previously acquired capabilities~\citep{biderman2024loralearnsforgets}. Non-parametric retrieval instead provides external evidence during generation. RAG~\citep{lewis2020rag} conditions on retrieved passages, and RETRO~\citep{borgeaud2022retro} incorporates retrieved chunks within the decoder. These methods provide access to corpus knowledge without first learning it in model parameters, but datastore search and evidence processing can add substantial inference latency~\citep{cao2025memorydecoder}.

Parametric memory learns from retrieved knowledge offline, enabling inference without explicit retrieval. MLP Memory~\citep{wei2026mlpmemory} and Memory Decoder~\citep{cao2025memorydecoder} learn to predict retrieval-derived next-token distributions. However, this inference advantage comes with substantial offline costs. Constructing token-level training targets requires retrieval at individual token positions across the corpus, resulting in substantial data construction costs~\citep{wei2026memorydecoderscale}. Training memory modules on these targets further incurs significant computational cost. Moreover, these methods rely on shallow fusion through output-probability interpolation, without incorporating memory into the decoder's hidden states. Intuitively, integrating memory into intermediate hidden states allows subsequent layers to further process and combine it with the evolving context. The broader challenge is to enable language models to acquire external knowledge efficiently and use it effectively.

In this work, we introduce \method{} (DFM), a parametric memory architecture that combines efficient knowledge acquisition with deep fusion within the decoder (Figure~\ref{fig:architecture}). At a high level, DFM shares memory supervision across multiple token predictions while allowing each token to use the learned memory according to its context. We jointly train a memory predictor and a memory fusion module, using retrieved target memory to supervise the predictor and the standard language modeling objective to train both modules. In this way, external knowledge is not only stored in the predictor's parameters, but also injected into intermediate hidden states through the fusion module, allowing subsequent decoder layers to utilize the memory during generation. The LLM backbone remains frozen throughout, and memory prediction is performed at the chunk level.

Our experiments demonstrate improvements in both efficiency and downstream performance. Compared with existing parametric memory approaches that construct token-level memory targets, DFM requires $32\times$ fewer retrieval queries and $2048\times$ fewer retrieved items to construct its training data. DFM achieves the highest average score across six QA tasks on both backbones, outperforming the strongest baseline by 12.71\% on Mistral-7B-v0.3 and 11.19\% on Qwen3-8B-Base. Beyond QA, adaptation across code, mathematics, and law yields average gains of 6.51 and 4.38 points over Qwen3-4B-Base and Qwen3-8B-Base, respectively. Ablations further highlight the importance of fusion placement and gate initialization, while showing comparatively small changes in average QA performance across the tested chunk sizes and numbers of memory slots.

Our contributions are summarized as follows:
\begingroup
\setlength{\leftmargini}{1pc}
\begin{itemize}
    \item We introduce \textbf{DFM, a parametric memory architecture with deep fusion}, which integrates learned corpus knowledge into intermediate hidden states so that subsequent decoder layers can process memory together with the evolving context. Inference requires no explicit retrieval.
    \item We reduce \textbf{the overhead of supervision construction and memory training} through chunk-level supervision and end-to-end training with a frozen backbone. For training data construction, DFM requires $32\times$ fewer retrieval queries and $2048\times$ fewer retrieved items than existing parametric memory approaches based on token-level targets.
    \item We demonstrate \textbf{improved QA performance and domain adaptation}. Across six QA tasks, DFM performs better over the strongest baseline by 12.71\% on Mistral-7B-v0.3 and 11.19\% on Qwen3-8B-Base. Across code, mathematics, and law, average gains over Qwen3-4B-Base and Qwen3-8B-Base reach 6.51 and 4.38 points, respectively.
\end{itemize}
\endgroup

%% file: sections/02_related_work.tex
\section{Related Work}
\label{sec:related-work}

\textbf{Model Adaptation.}
Continued pretraining adapts language models to additional corpora through further language modeling~\citep{gururangan2020dont}, while LoRA~\citep{hu2022lora} offers a parameter-efficient way to adapt the model.
Such adaptation can cause catastrophic forgetting of previously acquired capabilities~\citep{biderman2024loralearnsforgets}.
\method{} keeps the backbone frozen and learns an added memory module from retrieval-derived targets, jointly training memory prediction and fusion.

\textbf{Retrieval-Augmented Language Models.}
Non-parametric memory exposes corpus information through several interfaces.
$k$NN-LM retrieves context--token pairs and interpolates a neighbor-derived next-token distribution with the language model output~\citep{khandelwal2020knnlm}, whereas RAG conditions generation on retrieved passages~\citep{lewis2020rag}.
RETRO combines chunk retrieval with cross-attention inside the Transformer~\citep{borgeaud2022retro}.
xRAG projects a retrieved document embedding into a single token representation prepended to the query, reducing the context passed to the generator~\citep{cheng2024xrag}.
These methods retain an external retrieval dependency at inference.
\method{} predicts continuous memory parametrically and uses it inside the decoder without online search.

\textbf{Parametric Memory.}
MLP Memory trains an MLP to predict $k$NN token distributions from language model representations~\citep{wei2026mlpmemory}.
Memory Decoder learns a similar distributional target with a standalone decoder~\citep{cao2025memorydecoder}.
Memory Decoder at Scale extends this approach to large-scale general and domain memory pretraining~\citep{wei2026memorydecoderscale}.
MemSFT specializes frozen backbones through a domain memory and a learned token-wise router over the backbone and memory output distributions~\citep{wang2026memsft}.
These approaches avoid explicit retrieval at inference, but token-level retrieval incurs substantial data construction costs, and output-probability interpolation provides only shallow fusion.
\method{} reduces the cost of supervision construction by sharing target memory across tokens within each chunk and fuses predicted memory into hidden states through gated cross-attention.

%% file: sections/03_method.tex
\section{\method{}}
\label{sec:method}

\method{} consists of a causal chunk-level memory predictor and memory cross-attention modules inserted into selected decoder layers (Figure~\ref{fig:architecture}). We construct target memory offline, then jointly train the added modules with language-modeling and memory-prediction objectives while keeping the backbone frozen. At inference, the predictor supplies memory directly from the input sequence.

\input{sections/07_architecture}

\subsection{Training data construction}
\label{sec:training_data}

We partition each training sequence into non-overlapping chunks $X_0,\ldots,X_{J-1}$ of $C$ tokens. For each chunk $X_j$ with $j\geq 1$, an offline retriever encodes the preceding chunk $X_{j-1}$ and retrieves its top-$K$ neighbors from a corpus datastore. Each retrieved chunk is represented by a fixed dense embedding produced by a frozen text encoder. Stacking these embeddings in retrieval-rank order gives the \emph{target memory} $V_j\in\mathbb{R}^{K\times d_m}$, where $d_m$ is the embedding dimension. This ordering aligns the target memory with the predictor's output slots.

Each training example pairs the original token sequence with its chunk-level target memory. Each target memory supervises the predictor output shared across token predictions in $X_j$. Predictions for the first chunk use no memory, and this chunk is excluded from memory supervision because it has no preceding-chunk query. Target memory is prepared offline and remains fixed during model training.

\subsection{Training}
\label{sec:training}

\textbf{Memory Predictor.}
The predictor generates $K$ continuous memory vectors for each chunk from the preceding context (Figure~\ref{fig:architecture}(a)). Let $e_i=E_\theta(x_i)$ denote the embedding of token $x_i$ in the frozen backbone. For $j\geq1$, we mean-pool the token embeddings in the preceding chunk $X_{j-1}$ and, when needed, apply a learned linear projection to match the input width of the memory predictor. We denote the resulting chunk embedding by $C_j$. The predictor uses the prefix of these embeddings to produce memory for $X_j$:
\begin{equation}
 \widehat V_j=\operatorname{Predictor}_\phi(C_{\leq j}).
 \label{eq:continuous_prediction}
\end{equation}
The first input $C_0$ is a masked placeholder. The predictor combines a causal Transformer with $K$ separate linear output heads, one per memory slot. The \emph{predicted memory} $\widehat V_j\in\mathbb{R}^{K\times d_m}$ stacks the vectors from these heads in retrieval-rank order. Causal attention lets the prediction for $X_j$ use $C_1,\ldots,C_j$, which summarize chunks $X_0,\ldots,X_{j-1}$.

\textbf{Memory Fusion.}
Let $P_\psi$ denote the shared memory projector in Figure~\ref{fig:architecture}(a). It maps predicted memory into the decoder hidden space, yielding the \emph{projected memory} $M_j=P_\psi(\widehat V_j)\in\mathbb{R}^{K\times d}$, where $d$ is the hidden width. The projector applies LayerNorm followed by two linear layers with a GELU activation independently to each slot. The resulting $M_j$ is reused across the selected decoder layers.

Consider one selected decoder layer in Mistral or Qwen3 (Figure~\ref{fig:architecture}(c)), omitting the layer index for clarity. Let $z_t=\operatorname{Norm}(h_t)$ be the normalized token state used by self-attention. The schematic omits this normalization step. Let $j$ denote the chunk containing the next token predicted at position $t$. Let $o_{t,a}$ denote the scaled dot-product cross-attention output of head $a$, with queries projected from $z_t$ and keys and values from $M_j$. Each of the $H$ heads applies a softmax over the $K$ slots. A learned gate controls the memory contribution of each head:
\begin{equation}
 g_t=\sigma(W_g z_t+b_g),\qquad u_t=W_O\operatorname{Concat}_{a=1}^{H}\!\left(g_{t,a}\,o_{t,a}\right).
 \label{eq:memory_attention}
\end{equation}
The sigmoid gate $g_t\in(0,1)^H$ assigns one scalar to each token and head, shared across all channels in the head. It scales the head output before concatenation and output projection. Attention selects information from the shared memory for each token, while the gate controls how much each head contributes. For token predictions in the first chunk, the memory update is zero.

The memory branch runs in parallel with causal self-attention. Both outputs enter the same residual update:
\begin{equation}
 \widetilde h_t=h_t+\operatorname{SelfAttn}(z_{\leq t})+u_t.
 \label{eq:fusion_update}
\end{equation}
The updated state $\widetilde h_t$ then passes through the original decoder feed-forward sublayer, including the associated normalization and residual connection. The memory predictor and memory projector are shared across fusion layers. Each fusion layer uses layer-specific memory-attention and gate projections, separate from the self-attention parameters. Other decoder layers retain their original computation.

Memory alignment follows next-token prediction. With zero-based token positions, the state at $t$ predicts $x_{t+1}$ and uses chunk index $j=\lfloor(t+1)/C\rfloor$. At a chunk boundary, the preceding chunk is complete before the pooled query is used. Together with the memory predictor, this ensures that memory generation uses only the input available for the next-token prediction.

\textbf{Joint Training Objective.}
We jointly train the predictor and fusion modules end to end using next-token cross-entropy, while additionally supervising the memory predictor with a mean squared error loss between predicted and target memory:
\begin{equation}
 \mathcal{L}=\mathcal{L}_{\mathrm{LM}}+\beta\mathcal{L}_{\mathrm{MSE}},
 \qquad
 \mathcal{L}_{\mathrm{MSE}}=
 \frac{\sum_{j=1}^{J-1}\|\widehat V_j-V_j\|_F^2}
 {(J-1)Kd_m}.
 \label{eq:training_loss}
\end{equation}
The MSE excludes $X_0$, which has no preceding chunk. We set $\beta=1$. The MSE is computed directly against the L2-normalized target embeddings, without renormalizing the predictions. The language model consumes predicted memory throughout training. Target memory is used only for supervision. Gradients from $\mathcal{L}_{\mathrm{LM}}$ pass through memory fusion into the predictor, while $\mathcal{L}_{\mathrm{MSE}}$ directly supervises the predicted memory slots. The predictor, query projection, memory projector, cross-attention, and gate parameters are optimized jointly. The backbone, retrieval encoder, and target memory remain fixed.

\subsection{Inference}
\label{sec:inference}

At inference, the language model uses the same predictor and fusion modules without constructing target memory. Chunk queries are formed from backbone input embeddings, and the memory predictor supplies predicted memory to the selected decoder layers. No datastore search, retrieved text, or external embedding encoder is required. For QA evaluation, predicted memory is computed from the visible prompt and remains fixed during answer generation. Selected decoder layers attend to the predicted memory only at positions associated with the prompt chunks. As defined in Eq.~\ref{eq:memory_attention}, a sigmoid gate conditioned on the current token state scales each head before output projection. With layer-specific gate parameters, the contribution of the fixed memory can vary across tokens, heads, and layers.

%% file: sections/07_architecture.tex
\begin{figure}[t]
    \centering
    \includegraphics[width=\linewidth]{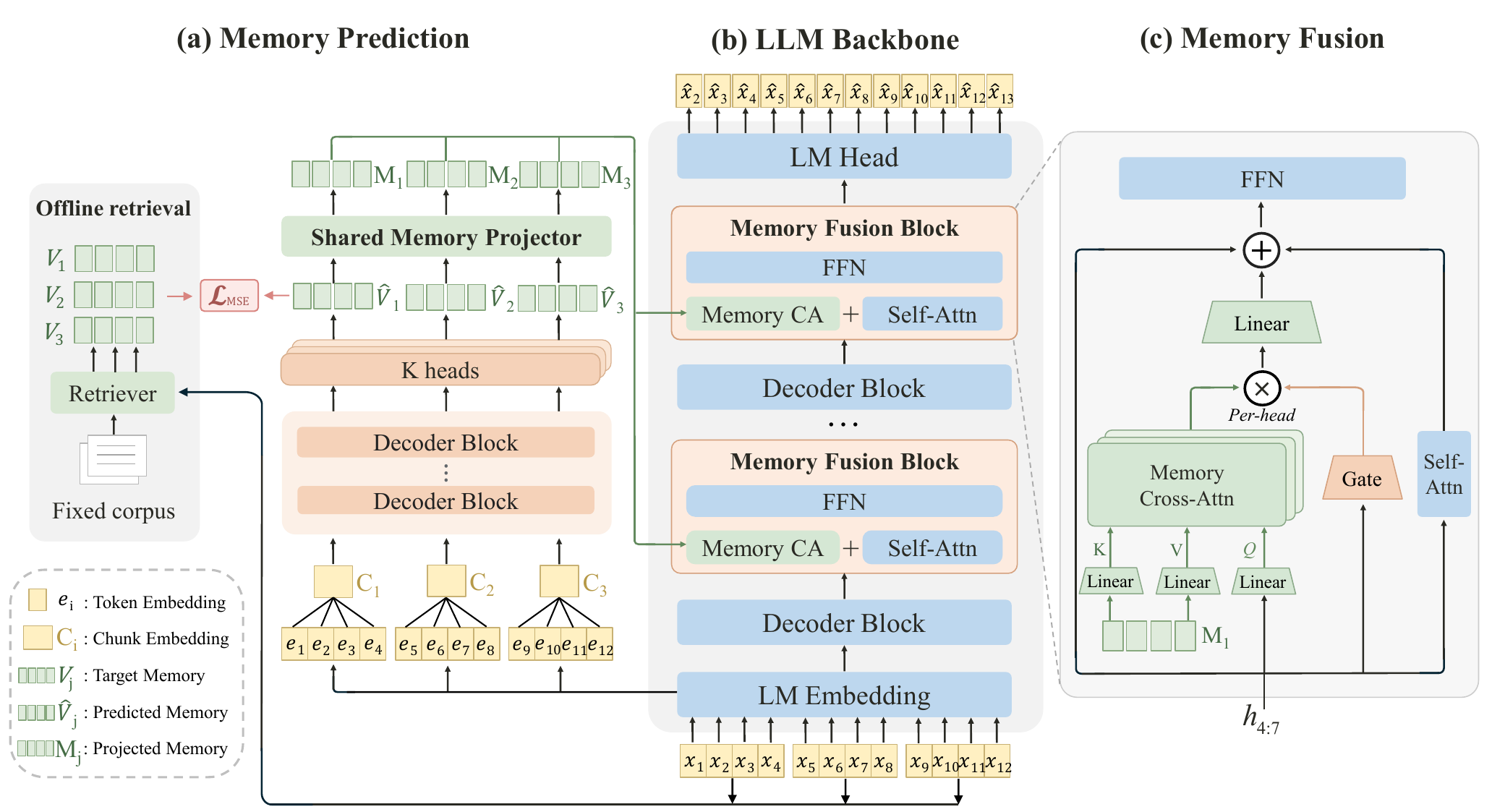}
    \caption{Overview of \method{}. \textbf{(a) Memory Prediction.} The memory predictor maps pooled chunk embeddings to predicted memory, supervised by offline target memory. A shared projector maps predicted memory to the decoder's hidden space. \textbf{(b) LLM Backbone.} Selected decoder layers incorporate memory for next-token prediction. \textbf{(c) Memory Fusion.} Memory predicted from the preceding context supports token predictions in the next chunk. Per-head gates control the memory cross-attention contribution alongside self-attention. Inference requires no retrieval.}
    \label{fig:architecture}
\end{figure}

%% file: sections/04_experiments.tex
\section{Experiments}
\label{sec:experiments}

We evaluate the effectiveness and efficiency of DFM across knowledge-intensive question answering and domain adaptation. We first describe the datasets, training configurations, and baselines (\S\ref{sec:experimental_setup}). We then evaluate DFM on six knowledge-intensive question-answering tasks (\S\ref{sec:qa_results}) and examine its effectiveness beyond QA through domain adaptation in Code, Math, and Law (\S\ref{sec:domain_injection}). Finally, we compare parameter overhead, training time, and target-construction workload with token-level memory (\S\ref{sec:efficiency}).

\subsection{Experimental Setup}
\label{sec:experimental_setup}

\textbf{Datasets.}
For question answering, we train on English Wikipedia from December 2018, comprising approximately 4B tokens. We evaluate single-hop QA on NQ~\citep{kwiatkowski2019nq} and WebQA~\citep{berant2013webquestions}, and multi-hop QA on HotpotQA~\citep{yang2018hotpotqa}, 2WikiMultiHopQA~\citep{ho2020twowiki}, MuSiQue~\citep{trivedi2022musique}, and Bamboogle~\citep{press2023bamboogle}. For domain adaptation, we sample subsets from Swallow-Code~\citep{fujii2025rewritingpretrainingdataboosts} and Swallow-Math~\citep{fujii2025rewritingpretrainingdataboosts}, and use legal text from DISC-Law-SFT~\citep{yue2023disclawllm}. We evaluate Code on HumanEval~\citep{chen2021humaneval} and HumanEval+~\citep{liu2023evalplus}, Math on GSM8K~\citep{cobbe2021gsm8k}, and Law on LawBench~\citep{fei2023lawbench}. Appendices~\ref{app:current_experiments} and~\ref{app:domain_details} give data-construction and evaluation details for QA and domain adaptation.

\textbf{Training Details.}
We use Mistral-7B-v0.3~\citep{jiang2023mistral} and Qwen3-8B-Base~\citep{yang2025qwen3} for question answering, and Qwen3-4B-Base and Qwen3-8B-Base~\citep{yang2025qwen3} for domain adaptation. DFM uses 32-token chunks and target memory formed from the top-16 chunk embeddings retrieved using Qwen3-Embedding-0.6B~\citep{zhang2025qwen3embedding}, with gated memory fusion at layers $[9,16,24,31]$. The added modules contain 0.75B parameters on both QA backbones and 0.6B on Qwen3-4B-Base. We freeze the backbone and train the added modules for one epoch with next-token cross-entropy and memory-prediction MSE, using AdamW with learning rate $1\times10^{-5}$. Wikipedia training uses 2,048-token sequences and a global batch size of 128 on 16 A800 80GB GPUs. For domain adaptation, we train a separate model for each domain with 262,144 tokens per optimizer update. Further architectural details and training configurations are given in Appendix~\ref{app:experimental_details}.

\textbf{Baselines.}
We compare against three non-parametric and three parametric baselines. Among non-parametric methods, \textbf{RAG}~\citep{lewis2020rag} augments the input with the top-1 passage retrieved using Qwen3-Embedding-0.6B. \textbf{xRAG}~\citep{cheng2024xrag} projects a retrieved document's dense embedding into a single token prepended to the question. \textbf{RETRO}~\citep{borgeaud2022retro} incorporates retrieved chunks through cross-attention. Among parametric methods, \textbf{CPT} updates all backbone parameters through continued pretraining for one epoch. \textbf{LoRA}~\citep{hu2022lora} trains low-rank adapters for one epoch, with its added parameter budget matched to DFM. \textbf{MLP Memory}~\citep{wei2026mlpmemory} learns to predict token-level kNN distributions from Wikipedia 2018 and interpolates its predictions with the backbone's output probabilities.

\input{tables/exp_main_qa}

\subsection{Question Answering Performance}
\label{sec:qa_results}

\textbf{Overall Performance.}
Table~\ref{tab:knowledge_injection} compares DFM with non-parametric and parametric methods across six QA benchmarks. DFM achieves the highest average on both backbones without online retrieval. On Mistral, it raises the six-task mean from xRAG's 24.14 to 27.21, a \textbf{12.71\% relative improvement} over the strongest baseline. On Qwen3-8B-Base, it improves from MLP Memory's 23.80 to 26.46, a \textbf{11.19\% relative improvement}. These correspond to absolute gains of 3.07 and 2.66 points, showing comparable relative benefits across the two backbone families. Relative to Base, DFM gains 13.59 and 3.75 points, improving all six tasks on Mistral and five on Qwen3. It also outperforms CPT and LoRA on every Mistral task and five Qwen3 tasks.

\textbf{Single-Hop Question Answering.}
On Mistral, DFM reaches 33.57 on NQ and 42.47 on WebQA, improving over Base by 13.27 and 13.19 points. These scores exceed the strongest baseline on each task by 6.18 and 2.26 points, showing gains beyond those obtained through either explicit retrieval or parameter adaptation alone. On Qwen3, NQ improves from 17.60 to 27.46, exceeding RAG by 3.04 points. WebQA instead decreases from 41.78 to 39.47, while MLP Memory achieves 43.26. The contrasting WebQA outcomes show that memory adaptation is not uniformly beneficial across backbones, even when the original weights remain frozen.

\textbf{Multi-Hop Question Answering.}
DFM improves over Base on all four multi-hop benchmarks for both backbones. On Mistral, it leads on HotpotQA, MuSiQue, and Bamboogle, with Bamboogle increasing from 3.91 to 22.65. Its 2Wiki score rises from 5.87 to 25.00, although RAG remains stronger at 31.02. On Qwen3, DFM achieves the best results on 2Wiki, MuSiQue, and Bamboogle, including a 7.55-point gain over Base on 2Wiki. It also improves HotpotQA to 26.46, below xRAG's 27.74. These results indicate that predicted memory can support tasks requiring multiple pieces of evidence, while explicit retrieval retains an advantage on some benchmarks.

\subsection{Domain Adaptation}
\label{sec:domain_injection}

\input{tables/exp_domain_transfer}

Table~\ref{tab:domain_transfer} examines whether DFM also supports reasoning-oriented tasks and adaptation to specialized domains. Across Code, Math, and Law, DFM improves all four benchmarks over Base on both Qwen3-4B-Base and Qwen3-8B-Base. It achieves the highest average among the compared methods, reaching 64.69 on 4B and 67.52 on 8B, gains of 6.51 and 4.38 points.

\textbf{Code and Mathematical Reasoning.}
On 4B, DFM leads on HumanEval, HumanEval+, and GSM8K, improving over Base by 7.47, 5.48, and 2.65 points, respectively. It exceeds CPT by 2.89 and 4.26 points on the two code benchmarks and LoRA by 0.68 points on GSM8K. On 8B, DFM achieves the highest HumanEval score of 73.02 and raises HumanEval+ from 59.15 to 64.02. LoRA remains stronger on HumanEval+ and GSM8K, where DFM gains 0.83 points over Base. DFM also matches or exceeds RAG on these benchmarks, tying only on 8B HumanEval+. Memory adaptation thus benefits reasoning-oriented tasks at both scales, although its advantage over other adaptation methods varies by benchmark.

\textbf{Legal Domain Adaptation.}
Training on legal text raises LawBench from 29.96 to 40.41 on 4B and from 38.45 to 46.61 on 8B. RAG scores 24.35 and 26.89, leaving DFM ahead by 16.06 and 19.72 points, respectively. On 4B, DFM exceeds LoRA by 0.98 points but remains 1.25 points below CPT. On 8B, it achieves the best score, exceeding CPT and LoRA by 1.52 and 0.88 points. Per-task LawBench results appear in Appendix~\ref{app:domain_results}, with domain-specific training configurations in Appendix~\ref{app:domain_details}.

\subsection{Efficiency Analysis}
\label{sec:efficiency}

Figure~\ref{fig:efficiency_comparison} examines the resource requirements of chunk-level parametric memory relative to token-level MLP Memory. We compare added model capacity, training time on 16 A800 GPUs, and the offline retrieval workload for constructing targets from a common 4B-token corpus.

\input{figures/efficiency_comparison}

\textbf{Parameter and Training Costs.}
As shown in Figure~\ref{fig:efficiency_comparison}(a), DFM adds 0.75B parameters, approximately 50\% fewer than the 1.5B Mistral MLP Memory configuration~\citep{wei2026mlpmemory}. Trained on 16 A800 GPUs with the same corpus and number of epochs, MLP Memory and DFM take approximately 64 and 24 hours, respectively, reducing training time by approximately 62.5\%. Figure~\ref{fig:efficiency_comparison}(b) compares these measured training costs. Appendix~\ref{app:efficiency_accounting} provides the timing details.

\textbf{Target Construction.}
Figure~\ref{fig:efficiency_comparison}(c) shows that, for a 4B-token corpus, token-level construction requires 4B retrieval queries, whereas 32-token chunks require 125M, a 96.88\% reduction. Each query produces one target entry. With Top-1024 for MLP Memory and Top-16 for DFM, the total number of retrieved items falls from $4\mathrm{B}\times1024=4.096\mathrm{T}$ to $125\mathrm{M}\times16=2\mathrm{B}$, a 99.95\% reduction. By reducing both the number of retrieval queries and the neighbors returned per query, chunk-level supervision substantially lowers the retrieval overhead of target construction.

\FloatBarrier

%% file: tables/exp_main_qa.tex
\begin{table}[t]
\caption{Question answering performance on two single-hop benchmarks and four multi-hop benchmarks. Avg is the unweighted mean across all six tasks. Red subscripts show decreases from Base. Blue highlights DFM. Bold scores indicate the best results within each backbone.}
\label{tab:knowledge_injection}
\setlength{\tabcolsep}{6pt}
\begin{center}
\begin{tabularx}{\textwidth}{@{}Xccccccc<{\hspace*{20pt}}@{}}
\toprule
\multirow{2}{*}[-.5\dimexpr\aboverulesep+\belowrulesep+\cmidrulewidth\relax]{Method} & \multicolumn{2}{c}{Single-hop QA} & \multicolumn{4}{c}{Multi-hop QA} & \multirow{2}{*}[-.5\dimexpr\aboverulesep+\belowrulesep+\cmidrulewidth\relax]{Avg} \\
\cmidrule(lr){2-3}\cmidrule(lr){4-7}
 & NQ & WebQA & HotpotQA & 2Wiki & MuSiQue & Bamboogle & \\
\midrule
\multicolumn{8}{@{}l}{\textbf{Mistral-7B-v0.3}} \\
Base & 20.30 & 29.28 & 20.96 & 5.87 & 1.43 & 3.91 & 13.63 \\
\rowcolor{dfmgroup}\multicolumn{8}{c}{\emph{Non-parametric methods}} \\
RAG & 24.03 & 27.90 & 29.09 & \textbf{31.02} & 9.87 & 18.56 & 23.41 \\
xRAG & 27.39 & 40.06 & 27.55 & 27.78 & 9.57 & 12.51 & 24.14 \\
RETRO & 20.26 & 29.48 & 21.06 & 5.90 & 1.48 & 4.19 & 13.73 \\
\rowcolor{dfmgroup}\multicolumn{8}{c}{\emph{Parametric methods}} \\
CPT & 11.44 & 32.92 & 12.13 & 14.22 & 4.51 & 5.65 & \scoredecrease{13.48}{\textminus{}0.15} \\
LoRA & 14.24 & 40.21 & 13.43 & 14.98 & 4.90 & 5.08 & 15.47 \\
MLP Memory & 26.32 & 39.57 & 24.83 & 22.36 & 7.36 & 12.08 & 22.09 \\
\rowcolor{dfmhighlight}\multicolumn{1}{@{}>{\columncolor{dfmhighlight}[0pt][\tabcolsep]}l}{DFM} & \textbf{33.57} & \textbf{42.47} & \textbf{29.53} & 25.00 & \textbf{10.05} & \textbf{22.65} & \multicolumn{1}{>{\columncolor{dfmhighlight}[\tabcolsep][0pt]}c<{\hspace*{20pt}}@{}}{\textbf{27.21}} \\
\midrule
\multicolumn{8}{@{}l}{\textbf{Qwen3-8B-Base}} \\
Base & 17.60 & 41.78 & 24.65 & 24.83 & 9.72 & 17.66 & 22.71 \\
\rowcolor{dfmgroup}\multicolumn{8}{c}{\emph{Non-parametric methods}} \\
RAG & 24.42 & 36.32 & 25.62 & 22.21 & 5.73 & 10.86 & \scoredecrease{20.86}{\textminus{}1.85} \\
xRAG & 24.34 & 41.54 & \textbf{27.74} & 22.46 & 9.67 & 15.30 & 23.51 \\
RETRO & 17.65 & 42.32 & 24.55 & 24.85 & 9.63 & 17.11 & \scoredecrease{22.69}{\textminus{}0.02} \\
\rowcolor{dfmgroup}\multicolumn{8}{c}{\emph{Parametric methods}} \\
CPT & 11.83 & 42.47 & 11.87 & 13.36 & 4.71 & 6.51 & \scoredecrease{15.13}{\textminus{}7.58} \\
LoRA & 11.72 & 40.16 & 11.67 & 13.41 & 4.59 & 6.43 & \scoredecrease{14.66}{\textminus{}8.04} \\
MLP Memory & 18.30 & \textbf{43.26} & 25.27 & 27.47 & 10.81 & 17.66 & 23.80 \\
\rowcolor{dfmhighlight}\multicolumn{1}{@{}>{\columncolor{dfmhighlight}[0pt][\tabcolsep]}l}{DFM} & \textbf{27.46} & 39.47 & 26.46 & \textbf{32.38} & \textbf{12.89} & \textbf{20.09} & \multicolumn{1}{>{\columncolor{dfmhighlight}[\tabcolsep][0pt]}c<{\hspace*{20pt}}@{}}{\textbf{26.46}} \\
\bottomrule
\end{tabularx}
\end{center}
\end{table}

%% file: tables/exp_domain_transfer.tex
\begin{table}[!htbp]
\caption{Domain adaptation on Qwen3-4B-Base and Qwen3-8B-Base across Code, Math, and Law. Bold and underlined scores mark the best and second-best results within each backbone. Red subscripts show percentage-point decreases from Base.}
\label{tab:domain_transfer}
\setlength{\tabcolsep}{2pt}
\begin{center}
\begin{tabularx}{0.86\textwidth}{@{}l>{\centering\arraybackslash}X>{\centering\arraybackslash}X>{\centering\arraybackslash}X>{\centering\arraybackslash}X>{\centering\arraybackslash}X@{}}
\toprule
\multirow{2}{*}[-.5\dimexpr\aboverulesep+\belowrulesep+\cmidrulewidth\relax]{Method} & \multicolumn{2}{c}{Code} & Math & Law & \multirow{2}{*}[-.5\dimexpr\aboverulesep+\belowrulesep+\cmidrulewidth\relax]{Avg} \\
\cmidrule(lr){2-3}\cmidrule(lr){4-4}\cmidrule(lr){5-5}
 & HumanEval & HumanEval+ & GSM8K & LawBench & \\
\midrule
\multicolumn{6}{@{}l}{\textbf{Qwen3-4B-Base}} \\
Base & 62.04 & 57.32 & 83.40 & 29.96 & 58.18 \\
RAG & 62.80 & \scoredecrease{55.49}{\textminus{}1.83} & 84.99 & \scoredecrease{24.35}{\textminus{}5.61} & \scoredecrease{56.91}{\textminus{}1.27} \\
CPT & \underline{66.62} & \underline{58.54} & 84.84 & \textbf{41.66} & \underline{62.92} \\
LoRA & \scoredecrease{59.45}{\textminus{}2.59} & \underline{58.54} & \underline{85.37} & 39.43 & 60.70 \\
\rowcolor{dfmhighlight}\multicolumn{1}{@{}>{\columncolor{dfmhighlight}[0pt][\tabcolsep]}l}{DFM} & \textbf{69.51} & \textbf{62.80} & \textbf{86.05} & \underline{40.41} & \multicolumn{1}{>{\columncolor{dfmhighlight}[\tabcolsep][0pt]}c@{}}{\textbf{64.69}} \\
\midrule
\multicolumn{6}{@{}l}{\textbf{Qwen3-8B-Base}} \\
Base & 69.36 & 59.15 & 85.60 & 38.45 & 63.14 \\
RAG & \underline{70.88} & \underline{64.02} & 85.60 & \scoredecrease{26.89}{\textminus{}11.56} & \scoredecrease{61.85}{\textminus{}1.29} \\
CPT & \scoredecrease{68.29}{\textminus{}1.07} & 59.76 & \underline{87.11} & 45.09 & 65.06 \\
LoRA & \scoredecrease{69.05}{\textminus{}0.31} & \textbf{65.85} & \textbf{87.19} & \underline{45.73} & \underline{66.96} \\
\rowcolor{dfmhighlight}\multicolumn{1}{@{}>{\columncolor{dfmhighlight}[0pt][\tabcolsep]}l}{DFM} & \textbf{73.02} & \underline{64.02} & 86.43 & \textbf{46.61} & \multicolumn{1}{>{\columncolor{dfmhighlight}[\tabcolsep][0pt]}c@{}}{\textbf{67.52}} \\
\bottomrule
\end{tabularx}
\end{center}
\end{table}

%% file: figures/efficiency_comparison.tex
\begin{figure}[!htb]
\centering
\includegraphics{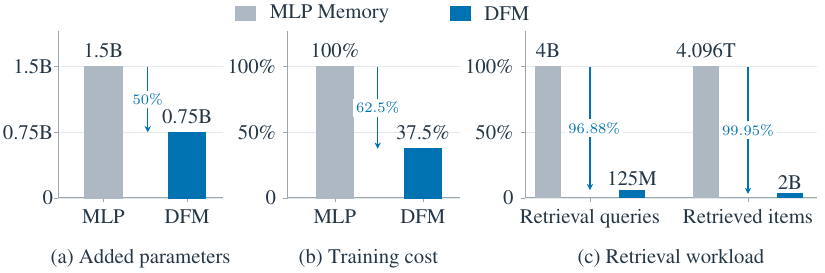}
\caption{Resource comparison with token-level MLP Memory. (a) Added parameters. (b) Training cost: MLP Memory and DFM take approximately 64 and 24 hours, respectively, on 16 A800 GPUs, a 62.5\% reduction. (c) Retrieval workload: retrieval queries and total retrieved items (queries $\times K$) for a 4B-token corpus. Reference bars in (b,c) equal 100\%; arrows show percentage reductions.}
\label{fig:efficiency_comparison}
\end{figure}

%% file: sections/05_analysis_and_ablations.tex
\section{Ablations and Analysis}
\label{sec:analysis_and_ablations}

We study chunk size, fusion-layer configurations, memory slots, and memory gating on Mistral-7B-v0.3 with predicted memory, varying one factor from the main configuration in Table~\ref{tab:knowledge_injection}. Figure~\ref{fig:ablation_structure} reports six-task QA averages. DFM is robust to the tested chunk sizes and slot counts, and to fusion configurations spanning intermediate and late layers. Appendix~\ref{app:ablations} gives exact settings, task-level results, and additional ablations on learning rate, projector sharing, predictor depth, and prediction loss. Appendix~\ref{app:case_studies} presents QA and offline-retrieval case studies.

\subsection{Ablation Studies}
\label{sec:analysis}

\input{figures/ablation_structure}

\textbf{Chunk Size.}
Figure~\ref{fig:ablation_structure}(a) reports averages of 26.51, 27.21, and 27.07 for chunk sizes 16, 32, and 64, respectively. The 32-token setting performs best. Doubling the chunk size to 64 halves target-construction queries with only a 0.14-point drop. The 16-token setting doubles retrieval queries without improving the mean, so finer-grained supervision is not necessarily more effective. This supports 32-token chunks as a practical balance between QA performance and construction workload.

\textbf{Fusion Depth and Placement.}
In Figure~\ref{fig:ablation_structure}(b), two to eight fusion layers starting at layer 9 achieve 27.09--27.54, versus 21.11 with final-layer-only fusion and 22.07 with the final four layers. With four fusion modules, distributing memory access across intermediate and late layers is more effective than concentrating it near the output. Four, six, and eight fusion layers remain within 0.13 points of one another, supporting the four-layer default without a consistent gain from greater fusion depth. The four-layer configuration $[2,10,21,31]$ scores 26.32, 0.89 points below the default. These results favor intermediate-to-late fusion, as also used in RETRO~\citep{borgeaud2022retro}. A possible explanation is that more developed semantic representations aid memory integration, with later layers processing the fused information.

\textbf{Number of Memory Slots.}
In Figure~\ref{fig:ablation_structure}(c), $K$ values of 2, 4, 8, and 16 yield averages of 26.88, 27.04, 27.09, and 27.21. Top-16 exceeds Top-8 by only 0.12 points: Top-8 scores higher on five tasks, while Top-16 improves Bamboogle. Even Top-2 remains within 0.33 points of Top-16, indicating that DFM maintains similar average QA performance with substantially fewer memory slots. This makes smaller $K$ a practical option for using fewer target vectors and reducing memory-attention computation while retaining similar average performance.

\subsection{Memory Gating}
\label{sec:memory_gating}

Table~\ref{tab:ablation_final_gate} examines the effect of token-wise, per-head gating on memory fusion and its sensitivity to initialization. Removing the gate reduces the six-task mean from 27.21 to 26.83, with the decrease driven mainly by Bamboogle and smaller, mixed changes on the other tasks. Initializing the gate bias at $0$ yields the highest mean, 27.90, and improves five of six tasks over the no-gate variant, providing further support for gated fusion. More negative initial biases yield lower means in this sweep. We retain $b_g=-2$ as a conservative default: with zero-initialized gate weights, $\sigma(-2)\approx0.119$ initially scales down the memory branch to limit its perturbation to the pretrained backbone, while allowing the gate to adapt during training.

\input{tables/exp_ablation_final_gate}
\FloatBarrier

%% file: figures/ablation_structure.tex
\begin{figure}[htbp]
\centering
\includegraphics{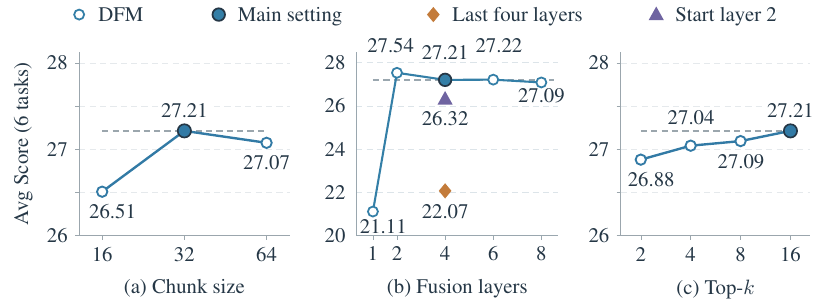}
\caption{DFM ablations with predicted memory on Mistral-7B-v0.3. Points show single-run six-task QA means. Filled circles and dashed lines mark the main setting (chunk 32, layers $[9,16,24,31]$, Top-16). In (b), the diamond and triangle use four fusion layers at $[28,29,30,31]$ and $[2,10,21,31]$, respectively.}
\label{fig:ablation_structure}
\end{figure}

%% file: tables/exp_ablation_final_gate.tex
\begin{table}[htbp]
\caption{Gate initialization and removal on Mistral-7B-v0.3. Gated variants change only the initial bias $b_g$; $b_g=-2$ is the default. Bold marks column maxima; Avg is the six-task mean.}
\label{tab:ablation_final_gate}
\begin{center}
\begin{tabular}{@{}lccccccc@{}}
\toprule
Gate & NQ & WebQA & HotpotQA & 2Wiki & MuSiQue & Bamboogle & Avg \\
\midrule
No gate & 34.08 & \textbf{42.72} & 29.80 & 24.94 & 10.08 & 19.35 & 26.83 \\
$b_g=-6$ & 32.34 & 42.18 & 28.54 & 23.55 & 9.03 & 14.94 & 25.10 \\
$b_g=-4$ & 32.70 & 42.22 & 29.39 & 25.98 & 10.42 & 19.70 & 26.73 \\
$b_g=-2$ & 33.57 & 42.47 & 29.53 & 25.00 & 10.05 & \textbf{22.65} & 27.21 \\
$b_g=0$ & \textbf{34.59} & 42.13 & \textbf{30.38} & \textbf{29.68} & \textbf{10.94} & 19.69 & \textbf{27.90} \\
\bottomrule
\end{tabular}
\end{center}
\end{table}

%% file: sections/05_conclusion.tex
\section{Conclusion}
\label{sec:conclusion}

We presented \method{} (DFM), a parametric memory architecture that combines efficient supervision construction with deep fusion within the decoder. Sharing target memory across tokens in each chunk reduces retrieval work, while gated cross-attention allows subsequent decoder layers to process predicted memory together with the evolving context. Memory prediction and fusion are trained end to end with the backbone frozen, and inference requires no explicit retrieval. DFM improves the six-task QA average over the strongest baseline by 12.71\% on Mistral-7B-v0.3 and 11.19\% on Qwen3-8B-Base. Beyond QA, DFM improves performance over the base models across code, mathematics, and law. These results demonstrate that shared supervision and learned internal fusion can support efficient knowledge acquisition and effective memory use.

%% file: sections/06_statements.tex
\section*{Acknowledgments}
This work is sponsored by the National Natural Science Foundation of China (NSFC) grant (No.
62576211) and the National Key Research and Development Program of China (No. 2023ZD0121402).
It is also the result of a collaborative project on novel language model architectures between Shanghai
Jiao Tong University (SJTU) and the Shanghai Artificial Intelligence Laboratory. The computational
resources required for pretraining the models were provided by the Shanghai AI Lab. This work is
also supported by the Specialized Program on Fundamental Research from Science and Technology
Commission of Shanghai Municipality (No. 2025SHZDZX025G09).

%% file: sections/08_appendix.tex
\input{sections/12_experimental_details}
\input{sections/13_additional_results}

\clearpage
\input{sections/14_ablation_studies}

\clearpage
\input{sections/15_efficiency_details}
\clearpage
\input{sections/16_additional_case_studies}

%% file: sections/12_experimental_details.tex
\section{Experimental Details}
\label{app:experimental_details}

\subsection{Question Answering}
\label{app:current_experiments}

\paragraph{Data construction.}
We use section and infobox text from Wikipedia December 2018. Records are grouped by normalized article titles and assigned to training, validation, and test splits in proportions of 99.8/0.1/0.1\% using seed 42. We remove exact text duplicates across splits, append EOS to each record, and pack records within each split into 2,048-token sequences. Incomplete final sequences are discarded. Tokenization yields 4.70B, 4.29B, and 4.08B training tokens for Mistral, Qwen3, and OLMo3, respectively. Each sequence is divided into non-overlapping 32-token chunks. We encode training chunks with the frozen Qwen3-Embedding-0.6B encoder, using last-token pooling and L2 normalization to obtain 1,024-dimensional representations. A FAISS IVF-PQ index supports offline neighbor retrieval. Following Section~\ref{sec:training_data}, each preceding chunk retrieves target memory for the next chunk. We retain the top-16 full chunk embeddings in retrieval-rank order, including the nearest neighbor, and mask the first chunk from memory supervision. The datastore contains training chunks only.

\paragraph{Training.}
DFM freezes the backbone and jointly trains its chunk-level memory predictor, shared projector, and gated fusion modules. The predictor has 28 layers, hidden width 1,024, FFN width 3,072, 16 query heads, and eight KV heads. Its 16 output heads predict the continuous memory slots. The L2-normalized targets have coordinates in $[-1,1]$, and the predictor clips output coordinates to this range. Memory cross-attention is inserted at zero-based layers $[9,16,24,31]$, with gate weights initialized to zero and gate biases to $-2$. OLMo3 retains its original attention-output normalization. The objective combines next-token cross-entropy and raw memory-prediction MSE with equal weights. We train for one epoch on 16 A800 80GB GPUs, with one sequence per GPU and eight gradient-accumulation steps, giving a global batch size of 128. AdamW uses learning rate $1\times10^{-5}$, $(\beta_1,\beta_2)=(0.9,0.999)$, $\epsilon=10^{-8}$, zero weight decay, and gradient clipping at 1. We use 500 warmup steps followed by cosine decay, BF16 precision, seed 42, and the final checkpoint for evaluation.

\paragraph{Baselines.}
\textbf{RAG}~\citep{lewis2020rag} retrieves the top-1 document from a Wikipedia training-corpus index using Qwen3-Embedding-0.6B and adds it to the question context. \textbf{xRAG}~\citep{cheng2024xrag} projects a retrieved document's dense embedding into a single token prepended to the question. \textbf{RETRO}~\citep{borgeaud2022retro} uses chunk-level retrieval and cross-attention to incorporate retrieved text into hidden states. \textbf{CPT} updates all backbone parameters for one epoch on Wikipedia at learning rate $2\times10^{-5}$. \textbf{LoRA}~\citep{hu2022lora} trains low-rank adapters for one epoch at the same learning rate of $2\times10^{-5}$, with the added parameter budget matched to DFM. \textbf{MLP Memory}~\citep{wei2026mlpmemory} learns to predict token-level kNN distributions from Wikipedia 2018 and interpolates its predictions with the backbone probabilities. As in MLP Memory, interpolation weights are task-specific; we select the best $\lambda\in\{0,0.05,\ldots,1\}$ on each reported evaluation split. In the task order of Table~\ref{tab:current_eval_protocol}, we use $\lambda=(0.55,0.55,0.55,1.00,0.90,1.00)$ for Mistral and $(0.35,0.45,0.25,0.80,0.55,0.00)$ for Qwen3.

\paragraph{Evaluation.}
Table~\ref{tab:current_eval_protocol} summarizes the six QA benchmarks. DFM uses zero-shot greedy decoding without online retrieval, gold documents, or chain-of-thought demonstrations. Questions are formatted with task-specific question--answer templates. Generation stops at a newline or the token limit, with period and comma as additional stop strings for WebQA. Evaluation batches contain one example per GPU, except WebQA, which uses four. For QA, each memory query is formed from the current visible prompt chunk rather than the preceding chunk used during training. The queries and predicted memory remain fixed during generation. For newly added chunk positions, the adapter pads queries and query-validity masks with zeros, so these positions receive no direct memory update. Generated answer chunks do not refresh the queries, and incremental predictor caching is not used in the reported implementation. F1 measures normalized token overlap and takes the maximum over accepted answers. Substring match checks whether a normalized reference answer occurs in the generated response. Scores are reported on a 0--100 scale, and Avg. is the unweighted mean across the six tasks.

\input{tables/exp_qa_evaluation}

\FloatBarrier
\subsection{Domain Adaptation}
\label{app:domain_details}

\paragraph{Data.}
Code and Math each use 100M training tokens sampled from Swallow-Code and Swallow-Math~\citep{fujii2025rewritingpretrainingdataboosts}, with document-level train/validation/test splits of 98/1/1\%. Law uses 123.5M training tokens from DISC-Law-SFT~\citep{yue2023disclawllm}, holding out 1\% of source-train for validation. We preserve its \texttt{Reference:} field when present, together with \texttt{Input:} and \texttt{Output:}. We deduplicate documents, screen benchmark overlap through normalized exact matching, append EOS, and pack sequences within each split. Target memory is constructed as in Appendix~\ref{app:current_experiments}, using training-only datastores, 32-token chunks, and Top-16 retrieval.

\paragraph{Training.}
We use the default DFM configuration for Qwen3-4B-Base and Qwen3-8B-Base, with the backbone frozen and the same memory predictor, shared projector, fusion layers, and gate initialization as in the QA experiments. DFM uses learning rate $1\times10^{-5}$, cosine decay, zero weight decay, and 11 warmup steps for Code/Math or 14 for Law. All adapted models train for one epoch in BF16 with 262,144 tokens per update. Sequence length is 8K for Code training on Qwen3-4B-Base and 2K for all other backbone--domain pairs. CPT and LoRA use the same data and sequence length as DFM. Both use learning rate $2\times10^{-5}$, with 3\% warmup and weight decay 0.1. LoRA uses rank 256, alpha 512, and zero dropout. We train a separate checkpoint for each domain.

\paragraph{RAG.}
RAG uses the frozen Base backbone without additional training. It retrieves the top-1 document from the corresponding domain's training corpus using Qwen3-Embedding-0.6B. Retrieved Code is prefixed as comments, while Math and Law use a background prefix. The original task prompt is preserved in full; only retrieved background is truncated when the input budget is exceeded. Few-shot examples, sample counts, generation settings, and scorers are identical to Base.

\paragraph{Evaluation.}
HumanEval uses three demonstrations and four samples per problem, while HumanEval+ uses zero demonstrations and one sample. Code reports sample-average execution correctness, estimating pass@1. GSM8K uses eight demonstrations and one sample. LawBench evaluates 500 examples for each of its 20 tasks in the zero-shot setting. All methods use a common \texttt{Input:}/\texttt{Output:} template without a chat template, greedy decoding, and the official task scorers. The LawBench score is the equal-weight macro average of the 20 task metrics. Main-table Avg. equally weights HumanEval, HumanEval+, GSM8K, and LawBench. Appendix~\ref{app:domain_results} provides the per-task LawBench results for the same models.

%% file: tables/exp_qa_evaluation.tex
\begin{table}[htbp]
\caption{Evaluation datasets and metrics for question answering. Max. tokens is the maximum number of newly generated tokens.}
\label{tab:current_eval_protocol}
\begin{center}
\begin{tabular*}{\textwidth}{@{\extracolsep{\fill}}llrcr@{}}
\toprule
Dataset & Split & Samples & Metric & Max. tokens \\
\midrule
NQ & Test & 3,610 & F1 & 15 \\
WebQA & Test & 2,032 & Substring match & 100 \\
\addlinespace
HotpotQA & Validation & 7,405 & F1 & 15 \\
2Wiki & Validation & 12,576 & F1 & 32 \\
MuSiQue & Validation & 2,417 & F1 & 32 \\
Bamboogle & Test & 125 & F1 & 32 \\
\bottomrule
\end{tabular*}
\end{center}
\end{table}

%% file: sections/13_additional_results.tex
\section{Additional Experimental Results}
\label{app:additional_results}

\subsection{Question Answering Performance}
\label{app:olmo3}

Table~\ref{tab:olmo3_qa} extends the Wikipedia experiment to OLMo-3-1025-7B under the same six-task protocol as Table~\ref{tab:knowledge_injection}. DFM increases the mean from 17.99 to 25.49, a gain of 7.50 points. It improves four tasks over Base, including 22.60 points on 2Wiki and 12.12 on Bamboogle, while WebQA and HotpotQA decrease by 1.28 and 0.63 points. Among the methods in Table~\ref{tab:olmo3_qa}, DFM achieves the highest NQ, 2Wiki, and MuSiQue scores. Its mean exceeds the strongest baseline, RETRO at 23.63, by 1.86 points.

\input{tables/exp_olmo3_qa}
\FloatBarrier
\subsection{Domain Adaptation}
\label{app:domain_results}

Table~\ref{tab:lawbench_task_definitions} maps LawBench task IDs to their names and definitions, grouped into legal knowledge memorization, understanding, and application. Tables~\ref{tab:domain_law_4b_tasks} and~\ref{tab:domain_law_8b_tasks} report the 20 LawBench subtasks for Qwen3-4B-Base and Qwen3-8B-Base, respectively, comparing Base, RAG, CPT, LoRA, and DFM. They use the same checkpoints and zero-shot evaluation protocol as Table~\ref{tab:domain_transfer}, with RAG using the frozen Base model. The Macro row reports the equal-weight average over the 20 task metrics, corresponding to the LawBench score in the main table. Appendix~\ref{app:domain_details} gives the training and evaluation settings.

\input{tables/lawbench_task_definitions}
\input{tables/exp_domain_law_4b_tasks}
\input{tables/exp_domain_law_8b_tasks}

%% file: tables/exp_olmo3_qa.tex
\begin{table}[!htbp]
\caption{Question answering performance on OLMo-3-1025-7B across two single-hop and four multi-hop benchmarks. Avg is the six-task mean. Red subscripts show decreases from Base. Blue highlights DFM. Bold and underlined scores mark the best and second-best results.}
\label{tab:olmo3_qa}
\setlength{\tabcolsep}{2pt}
\begin{center}
\begin{tabularx}{\textwidth}{@{}Xcc<{\hspace*{14pt}}ccccc<{\hspace*{20pt}}@{}}
\toprule
\multirow{2}{*}[-.5\dimexpr\aboverulesep+\belowrulesep+\cmidrulewidth\relax]{Method} & \multicolumn{2}{c}{Single-hop QA\hspace*{14pt}} & \multicolumn{4}{c}{Multi-hop QA} & \multirow{2}{*}[-.5\dimexpr\aboverulesep+\belowrulesep+\cmidrulewidth\relax]{Avg} \\
\cmidrule(lr{19pt}){2-3}\cmidrule(lr){4-7}
 & NQ & WebQA & HotpotQA & 2Wiki & MuSiQue & Bamboogle & \\
\midrule
\multicolumn{8}{@{}l}{\textbf{OLMo-3-1025-7B}} \\
Base & 21.69 & \textbf{40.31} & \textbf{27.99} & 11.05 & 4.67 & 2.25 & 17.99 \\
RAG & 22.57 & 31.50 & 21.97 & 7.91 & 2.58 & 2.75 & \scoredecrease{14.88}{\textminus{}3.11} \\
xRAG & \underline{23.64} & 34.79 & 26.33 & 23.15 & \underline{11.95} & \textbf{16.65} & 22.75 \\
RETRO & 21.61 & 38.09 & \underline{27.97} & \underline{29.90} & 11.61 & 12.57 & \underline{23.63} \\
LoRA & 7.39 & 34.40 & 7.32 & 6.73 & 2.55 & 1.21 & \scoredecrease{9.93}{\textminus{}8.06} \\
\rowcolor{dfmhighlight}DFM & \textbf{26.08} & \underline{39.03} & 27.36 & \textbf{33.65} & \textbf{12.44} & \underline{14.37} & \textbf{25.49} \\
\bottomrule
\end{tabularx}
\end{center}
\end{table}

%% file: tables/lawbench_task_definitions.tex
\begin{table}[htbp]
\caption{LawBench task IDs and brief definitions, following \citet{fei2023lawbench}. The first digit identifies the legal knowledge category. These IDs are used in Tables~\ref{tab:domain_law_4b_tasks} and~\ref{tab:domain_law_8b_tasks}.}
\label{tab:lawbench_task_definitions}
\begin{center}
\begin{tabularx}{\textwidth}{@{}l>{\RaggedRight\arraybackslash}p{0.34\textwidth}>{\RaggedRight\arraybackslash}X@{}}
\toprule
ID & Task & Description \\
\midrule
\multicolumn{3}{@{}l}{\textit{Legal Knowledge Memorization}} \\
1-1 & Article Recitation & Recall a specified law article. \\
1-2 & Knowledge Question Answering & Answer a legal multiple-choice question. \\
\midrule
\multicolumn{3}{@{}l}{\textit{Legal Knowledge Understanding}} \\
2-1 & Document Proofreading & Correct errors in a legal sentence. \\
2-2 & Dispute Focus Identification & Identify issues disputed by the parties. \\
2-3 & Marital Disputes Identification & Classify types of marital disputes. \\
2-4 & Issue Topic Identification & Classify a legal inquiry by topic. \\
2-5 & Reading Comprehension & Extract answers from judgment text. \\
2-6 & Named-Entity Recognition & Extract legal entities of specified types. \\
2-7 & Opinion Summarization & Summarize news on a legal matter. \\
2-8 & Argument Mining & Select an argument opposing the plaintiff. \\
2-9 & Event Detection & Detect events mentioned in legal text. \\
2-10 & Trigger Word Extraction & Extract trigger words for given events. \\
\midrule
\multicolumn{3}{@{}l}{\textit{Legal Knowledge Application}} \\
3-1 & Fact-based Article Prediction & Predict article numbers from case facts. \\
3-2 & Scene-based Article Prediction & Generate article text for a legal scenario. \\
3-3 & Charge Prediction & Predict charges from facts and article IDs. \\
3-4 & Prison Term Prediction without Article Text & Use facts, charges, and article numbers. \\
3-5 & Prison Term Prediction with Article Text & Also provide the applicable article text. \\
3-6 & Case Analysis & Answer a multiple-choice case question. \\
3-7 & Criminal Damages Calculation & Calculate the amount involved in a crime. \\
3-8 & Consultation & Answer a request for legal advice. \\
\bottomrule
\end{tabularx}
\end{center}
\end{table}

%% file: tables/exp_domain_law_4b_tasks.tex
\begin{table}[htbp]
\caption{LawBench task results on Qwen3-4B-Base under zero-shot evaluation. DFM uses the same checkpoint as the main table. Task definitions appear in Table~\ref{tab:lawbench_task_definitions}. Bold marks the highest score among the displayed methods. Macro is the equal-weight average over all 20 task metrics.}
\label{tab:domain_law_4b_tasks}
\begin{center}
\begin{tabular*}{\textwidth}{@{\extracolsep{\fill}}lrrrrr@{}}
\toprule
Task & Base & RAG & CPT & LoRA & DFM \\
\midrule
1-1 & 18.76 & 11.41 & \textbf{42.46} & 31.70 & 22.23 \\
1-2 & 6.20 & 20.00 & 27.00 & 28.20 & \textbf{31.60} \\
2-1 & 16.83 & 20.06 & 26.82 & \textbf{35.36} & 24.13 \\
2-2 & 12.20 & 0.60 & 31.40 & 23.80 & \textbf{35.60} \\
2-3 & 32.73 & 16.10 & 32.75 & 30.16 & \textbf{38.97} \\
2-4 & 0.00 & 0.20 & \textbf{34.00} & 33.40 & 30.80 \\
2-5 & 6.02 & 3.29 & 38.35 & 41.46 & \textbf{54.54} \\
2-6 & \textbf{39.56} & 34.64 & 10.32 & 5.30 & 27.81 \\
2-7 & 38.15 & 14.43 & \textbf{49.34} & 48.68 & 40.85 \\
2-8 & 0.00 & 0.00 & 34.20 & \textbf{39.80} & 36.80 \\
2-9 & 35.64 & 14.89 & \textbf{63.71} & 63.32 & 57.32 \\
2-10 & 13.25 & 3.03 & 17.35 & 22.07 & \textbf{24.07} \\
3-1 & 50.22 & 44.65 & \textbf{72.03} & 68.87 & 37.32 \\
3-2 & 24.92 & 9.20 & 34.54 & 26.30 & \textbf{35.24} \\
3-3 & 46.11 & 34.16 & \textbf{51.14} & 48.00 & 44.96 \\
3-4 & \textbf{82.88} & 80.45 & 76.91 & 59.92 & 74.23 \\
3-5 & \textbf{83.02} & 81.31 & 76.72 & 65.50 & 76.12 \\
3-6 & 2.00 & 9.20 & 31.20 & \textbf{31.60} & 31.40 \\
3-7 & 72.00 & \textbf{75.20} & 65.80 & 68.00 & 66.20 \\
3-8 & \textbf{18.79} & 14.11 & 17.23 & 17.08 & 17.94 \\
\midrule
Macro & 29.96 & 24.35 & \textbf{41.66} & 39.43 & 40.41 \\
\bottomrule
\end{tabular*}
\end{center}
\end{table}

%% file: tables/exp_domain_law_8b_tasks.tex
\begin{table}[htbp]
\caption{LawBench task results on Qwen3-8B-Base under zero-shot evaluation. DFM uses the same checkpoint as the main table. Task definitions appear in Table~\ref{tab:lawbench_task_definitions}. Bold marks the highest score among the displayed methods. Macro is the equal-weight average over all 20 task metrics.}
\label{tab:domain_law_8b_tasks}
\begin{center}
\begin{tabular*}{\textwidth}{@{\extracolsep{\fill}}lrrrrr@{}}
\toprule
Task & Base & RAG & CPT & LoRA & DFM \\
\midrule
1-1 & 16.72 & 8.99 & \textbf{54.57} & 54.15 & 29.71 \\
1-2 & \textbf{70.60} & 17.20 & 32.20 & 30.00 & 53.80 \\
2-1 & 20.74 & 21.89 & \textbf{38.60} & 27.72 & 30.85 \\
2-2 & 0.20 & 0.40 & 14.60 & 9.60 & \textbf{36.80} \\
2-3 & 16.01 & 16.41 & 46.21 & 40.52 & \textbf{52.40} \\
2-4 & 0.00 & 0.00 & 33.80 & \textbf{35.00} & 34.40 \\
2-5 & \textbf{52.28} & 5.95 & 37.58 & 35.82 & 51.40 \\
2-6 & \textbf{44.52} & 40.11 & 2.22 & 36.87 & 33.50 \\
2-7 & 29.31 & 15.00 & 47.44 & \textbf{49.19} & 43.26 \\
2-8 & \textbf{41.40} & 1.80 & 33.80 & 37.00 & 37.00 \\
2-9 & 26.38 & 15.26 & \textbf{74.02} & 70.16 & 49.72 \\
2-10 & 17.01 & 3.96 & 32.16 & \textbf{32.19} & 25.02 \\
3-1 & 53.92 & 54.20 & 78.37 & \textbf{81.09} & 62.09 \\
3-2 & 12.86 & 9.66 & \textbf{40.65} & 39.85 & 38.44 \\
3-3 & \textbf{50.92} & 46.93 & 48.75 & 50.77 & 48.05 \\
3-4 & 81.95 & \textbf{84.73} & 81.77 & 81.70 & 81.60 \\
3-5 & 81.55 & \textbf{82.94} & 81.92 & 82.63 & 81.47 \\
3-6 & \textbf{58.40} & 22.00 & 33.60 & 31.80 & 52.60 \\
3-7 & 74.00 & \textbf{75.80} & 69.20 & 71.40 & 69.40 \\
3-8 & 20.16 & 14.63 & 20.42 & 17.09 & \textbf{20.65} \\
\midrule
Macro & 38.45 & 26.89 & 45.09 & 45.73 & \textbf{46.61} \\
\bottomrule
\end{tabular*}
\end{center}
\end{table}

%% file: sections/14_ablation_studies.tex
\section{Ablation Studies}
\label{app:ablations}

All ablations start from the Mistral-7B-v0.3 configuration in Table~\ref{tab:knowledge_injection} and change only the factor under study. They use predicted memory, a frozen backbone, one training epoch, and the same six-task QA protocol. All results are single runs.

\subsection{Chunk Size, Fusion Placement, and Memory Slots}
\label{app:ablation_structure}

\paragraph{Reference configuration.}
The reference for Figure~\ref{fig:ablation_structure} uses continuous predicted memory, 32-token chunks, target memory from Top-16 retrieval, a shared projector, gate bias $-2$, and learning rate $1\times10^{-5}$. Its fusion layers are $[9,16,24,31]$ (zero-based), and its six task scores match Table~\ref{tab:knowledge_injection} exactly. These structural sweeps retain the other named architectural settings. Appendix~\ref{app:current_experiments} gives the reference training recipe.

\input{figures/ablation_task_deltas}

\paragraph{Chunk size.}
Relative to 32 tokens, the 16-token setting improves 2Wiki and MuSiQue but decreases the other four scores. The 64-token setting improves 2Wiki by 1.03 points, with decreases of 0.03--1.08 on the other tasks. For a fixed token budget, target-construction query counts scale inversely with chunk size. This establishes a workload trade-off but does not measure elapsed construction time.

\paragraph{Fusion depth and placement.}
The one-, two-, four-, six-, and eight-layer configurations use $[31]$, $[9,31]$, $[9,16,24,31]$, $[9,13,18,22,27,31]$, and $[9,12,15,18,22,25,28,31]$, respectively. This sweep changes both layer count and placement. The two-layer setting attains the highest mean (27.54), while the four-layer reference is retained across experiments for consistency. The additional four-layer condition uses $[28,29,30,31]$ and scores 22.07. All six tasks decrease relative to the distributed four-layer reference, including 10.63 points on NQ and 7.84 on Bamboogle. An additional four-layer configuration, $[2,10,21,31]$, obtains 26.32, 0.89 points below the reference. It keeps the number of modules fixed but changes multiple fusion positions, so the difference cannot be attributed solely to the starting layer.

\paragraph{Memory slots.}
Figure~\ref{fig:ablation_task_deltas} shows opposing task-level changes. Top-8 exceeds Top-16 by 0.07--0.80 points on five tasks but scores 3.07 points lower on Bamboogle, which contains 125 examples. Across Top-2 to Top-16, the six-task mean varies by only 0.33 points, indicating robustness to the tested memory-slot counts.

\FloatBarrier
\subsection{Learning Rate}
\label{app:ablation_lr}

Table~\ref{tab:ablation_lr} compares the default learning rate $1\times10^{-5}$ with $2\times10^{-5}$, retaining 32-token chunks, Top-16, a shared projector, gate bias $-2$, and fusion layers $[9,16,24,31]$. The default obtains 27.21 versus 27.10, with higher scores on NQ, WebQA, HotpotQA, and Bamboogle but lower scores on 2Wiki and MuSiQue. The 0.11-point difference indicates that average QA performance is robust to this twofold change in learning rate.

\input{tables/exp_ablation_lr}

\FloatBarrier
\input{sections/14_final_component_ablations}

%% file: figures/ablation_task_deltas.tex
\begin{figure}[htbp]
\centering
\includegraphics{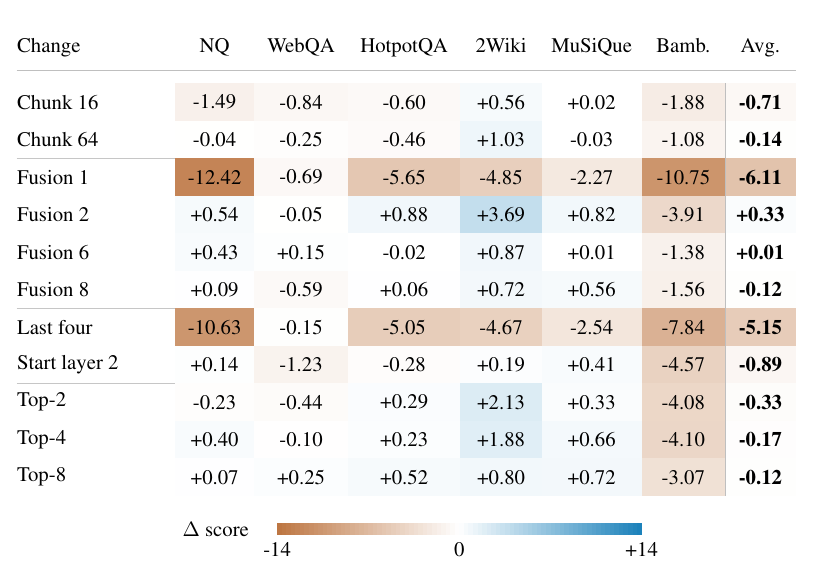}
\caption{Task-level score changes relative to the main DFM configuration on Mistral-7B-v0.3. Blue and brown indicate increases and decreases, respectively. Avg. gives the change in the unweighted six-task mean. Fusion labels denote the number of fusion layers, while Last four and Start layer 2 use $[28,29,30,31]$ and $[2,10,21,31]$, respectively. All differences are computed before rounding.}
\label{fig:ablation_task_deltas}
\end{figure}

%% file: tables/exp_ablation_lr.tex
\begin{table}[htbp]
\caption{Learning-rate ablation from the main DFM configuration. All other settings are fixed. The blue $1\times10^{-5}$ row is the default and reproduces the DFM results in Table~\ref{tab:knowledge_injection}. Bold marks the higher Avg.}
\label{tab:ablation_lr}
\begin{center}
\begin{tabularx}{\textwidth}{@{}Xccccccc@{}}
\toprule
Learning rate & NQ & WebQA & HotpotQA & 2Wiki & MuSiQue & Bamboogle & Avg \\
\midrule
$2\times10^{-5}$ & 33.47 & 41.93 & 29.18 & 28.60 & 10.43 & 19.00 & 27.10 \\
\rowcolor{dfmhighlight}$1\times10^{-5}$ & 33.57 & 42.47 & 29.53 & 25.00 & 10.05 & 22.65 & \textbf{27.21} \\
\bottomrule
\end{tabularx}
\end{center}
\end{table}

%% file: sections/14_final_component_ablations.tex

\subsection{Projector Sharing}
\label{app:ablation_final_projector}

Table~\ref{tab:ablation_final_projector} compares the shared memory projector with an independent projector at each fusion layer. The shared design scores 27.21 versus 26.88. Per-layer projectors improve WebQA, HotpotQA, 2Wiki, and MuSiQue but reduce NQ by 0.59 points and Bamboogle by 3.95 points. Sharing avoids replicating projector parameters across fusion layers and attains a slightly higher mean here.

\input{tables/exp_ablation_final_projector}
\FloatBarrier

\subsection{Memory Predictor Depth}
\label{app:ablation_final_predictor_depth}

Table~\ref{tab:ablation_final_predictor_depth} compares 14-, 28-, and 56-layer memory predictors while retaining the other training settings. Their six-task averages are 27.31, 27.21, and 26.47, respectively. The 0.10-point difference between 14 and 28 layers shows that average QA performance is stable when predictor depth is halved. Doubling depth reduces the mean by 0.75 points, driven mainly by a 4.74-point decrease on Bamboogle; NQ, HotpotQA, 2Wiki, and MuSiQue improve slightly. We retain 28 layers as the default for consistency across the main experiments.

\input{tables/exp_ablation_final_predictor_depth}
\FloatBarrier

\subsection{Memory Prediction Objective}
\label{app:ablation_final_prediction_loss}

Table~\ref{tab:ablation_final_prediction_loss} replaces raw MSE with a cosine-based memory prediction loss while retaining the other training settings. We compute one minus the cosine similarity for each predicted--target slot pair and average over all slots in the valid chunks, retaining $\beta=1$ in Eq.~\ref{eq:training_loss}. Raw MSE scores 27.21, compared with 26.87 for cosine loss. Cosine loss improves WebQA and 2Wiki by 0.39 and 0.48 points, respectively, but decreases the other four scores, with the largest drop on Bamboogle (2.23 points). The two objectives yield similar averages, with the default raw MSE higher by 0.34 points.

\input{tables/exp_ablation_final_prediction_loss}
\FloatBarrier

%% file: tables/exp_ablation_final_projector.tex
\begin{table}[htbp]
\caption{Shared versus per-layer memory projectors on Mistral-7B-v0.3. All other training settings are fixed. Blue marks the default shared projector. Bold marks the higher Avg. Avg is the six-task mean.}
\label{tab:ablation_final_projector}
\begin{center}
\begin{tabularx}{\textwidth}{@{}Xccccccc@{}}
\toprule
Projector & NQ & WebQA & HotpotQA & 2Wiki & MuSiQue & Bamboogle & Avg \\
\midrule
Per-layer & 32.98 & 43.21 & 29.88 & 26.35 & 10.18 & 18.70 & 26.88 \\
\rowcolor{dfmhighlight}Shared & 33.57 & 42.47 & 29.53 & 25.00 & 10.05 & 22.65 & \textbf{27.21} \\
\bottomrule
\end{tabularx}
\end{center}
\end{table}

%% file: tables/exp_ablation_final_predictor_depth.tex
\begin{table}[htbp]
\caption{Memory predictor depth on Mistral-7B-v0.3. Only the number of predictor layers changes. Blue marks the default 28-layer predictor. Bold marks the highest Avg. Avg is the six-task mean.}
\label{tab:ablation_final_predictor_depth}
\begin{center}
\begin{tabularx}{\textwidth}{@{}Xccccccc@{}}
\toprule
Predictor & NQ & WebQA & HotpotQA & 2Wiki & MuSiQue & Bamboogle & Avg \\
\midrule
14 layers & 34.27 & 41.83 & 30.14 & 26.31 & 10.30 & 21.03 & \textbf{27.31} \\
56 layers & 33.89 & 42.18 & 29.58 & 25.10 & 10.14 & 17.91 & 26.47 \\
\rowcolor{dfmhighlight}28 layers & 33.57 & 42.47 & 29.53 & 25.00 & 10.05 & 22.65 & 27.21 \\
\bottomrule
\end{tabularx}
\end{center}
\end{table}

%% file: tables/exp_ablation_final_prediction_loss.tex
\begin{table}[htbp]
\caption{Memory prediction objectives on Mistral-7B-v0.3. Only the memory prediction loss changes; all other training settings are fixed. Blue marks the default raw MSE objective. Bold marks the higher Avg. Avg is the six-task mean.}
\label{tab:ablation_final_prediction_loss}
\begin{center}
\begin{tabularx}{\textwidth}{@{}Xccccccc@{}}
\toprule
Objective & NQ & WebQA & HotpotQA & 2Wiki & MuSiQue & Bamboogle & Avg \\
\midrule
Cosine & 33.02 & 42.86 & 29.40 & 25.49 & 10.04 & 20.42 & 26.87 \\
\rowcolor{dfmhighlight}Raw MSE & 33.57 & 42.47 & 29.53 & 25.00 & 10.05 & 22.65 & \textbf{27.21} \\
\bottomrule
\end{tabularx}
\end{center}
\end{table}

%% file: sections/15_efficiency_details.tex
\section{Efficiency Analysis Details}
\label{app:efficiency_accounting}

Table~\ref{tab:efficiency_accounting} lists the numerical inputs and quantities used in Figure~\ref{fig:efficiency_comparison}. Training costs compare measured training times on 16 A800 GPUs for both methods; retrieval counts use a common 4B-token corpus.

\input{tables/exp_efficiency_accounting}

\paragraph{Reference configurations.}
We use the 1.5B MLP Memory configuration~\citep{wei2026mlpmemory} throughout the experiments, including the parameter and training-time comparisons. DFM adds approximately 0.75B parameters, a reduction of approximately 50\%.

\paragraph{Measured training time.}
Both methods are trained on 16 A800 GPUs using the same training corpus and number of epochs. Timings cover training only, excluding target construction and evaluation. MLP Memory takes approximately 64 hours, while DFM takes approximately 24 hours, corresponding to 1,024 and 384 GPU-hours. DFM therefore uses $24/64=37.5\%$ of the training time, a reduction of approximately 62.5\%. All reported costs use these rounded training times.

\paragraph{Construction counts.}
For $N=4\times10^9$ tokens, ignoring incomplete chunks and sequence-start masking, token-level targets require $N$ queries and chunk-level targets require $N/32$. Retrieving 1,024 versus 16 neighbors yields $1024N$ versus $16N/32$ returned slots. Their ratios are $32$ for queries and $2048$ for returned slots. The latter counts search outputs before MLP Memory removes the self neighbor and aggregates token probabilities. It does not count distinct tokens, search candidates examined, stored bytes, or FLOPs: DFM's continuous vector targets differ from token-level distribution targets.

%% file: tables/exp_efficiency_accounting.tex
\begin{table}[htbp]
\caption{Numerical accounting for the comparison in Figure~\ref{fig:efficiency_comparison}. Construction counts assume a common 4B-token corpus. Training hours are rounded measurements on 16 A800 GPUs for both methods.}
\label{tab:efficiency_accounting}
\begin{center}
\begin{tabularx}{\textwidth}{@{}Xrrl@{}}
\toprule
Quantity & MLP Memory & DFM & Basis \\
\midrule
Added parameters (B) & 1.50 & 0.75 & Configuration \\
A800 GPUs & 16 & 16 & Configuration \\
Measured training hours & 64 & 24 & Timing \\
Training GPU-hours & 1,024 & 384 & GPUs $\times$ hours \\
Target queries & 4B & 125M & Workload count \\
Neighbors per query & 1,024 & 16 & Configuration \\
Returned neighbor slots & 4.096T & 2B & Workload count \\
\bottomrule
\end{tabularx}
\end{center}
\end{table}

%% file: sections/16_additional_case_studies.tex
\section{Case Studies}
\label{app:case_studies}

\subsection{Question Answering}
\label{app:qa_cases}
\label{sec:case_study}

Figure~\ref{fig:dfm-qa-cases} compares answers from Base, MLP Memory, and DFM on six selected HotpotQA examples using Mistral-7B-v0.3. DFM answers all six correctly, while Base and MLP Memory give the same incorrect answer in each case. DFM identifies Manglietia from the stated species count and geographic origin, whereas both baselines choose Abronia. For the film comparison, DFM correctly selects \emph{What Ever Happened to Baby Jane?}, while both baselines select \emph{All the Marbles}. The other four cases cover a shared property of two bands, a composer's nationality, a university's association with a cartoon character, and the film underlying a Pixar short.

Cases were selected for a correct normalized exact match from DFM and zero exact match, substring match, and F1 from both baselines. MLP Memory uses $\lambda=0.55$. These selected successes illustrate differences in final answers, without establishing their frequency or revealing the models' intermediate reasoning.

\input{figures/dfm_qa_cases_main}
\clearpage

\subsection{Offline Retrieval Supervision}
\label{app:retrieval_cases}

Figure~\ref{fig:dfm-retrieval-supervision} presents seven examples of the offline retrieval used to construct training targets. Each row pairs a preceding 32-token query chunk with retrieved chunk text and the ground-truth next chunk. For Pakenham, the retrieved text supplies the network operator, Metro Trains Melbourne. For Bhutan cricket, it supplies the 2001 ICC membership and international debut. The other cases cover a chart position, railway privatization, Welsh place names, a mountain range, and a legal proceeding. Highlighted spans identify continuation-relevant information that is missing or incomplete in the query chunk.

The selected records exclude the query chunk itself, exact query-text matches, and chunks from the same source block. Cross-block near duplicates may remain. These examples illustrate the relevance of retrieved supervision targets. They do not directly measure the fidelity of predicted memory or attribute the answer differences above to particular retrieved chunks.

\input{figures/dfm_retrieval_cases_all7_warm}
\FloatBarrier

%% file: figures/dfm_qa_cases_main.tex
\begin{figure}[!htbp]
  \centering
  \includegraphics[width=\textwidth]{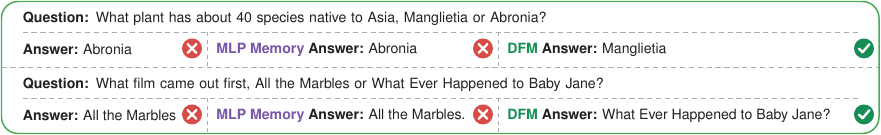}
  \par\medskip
  \includegraphics[width=\textwidth]{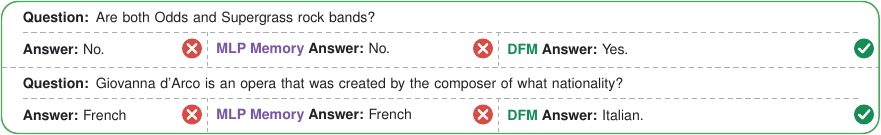}
  \par\medskip
  \includegraphics[width=\textwidth]{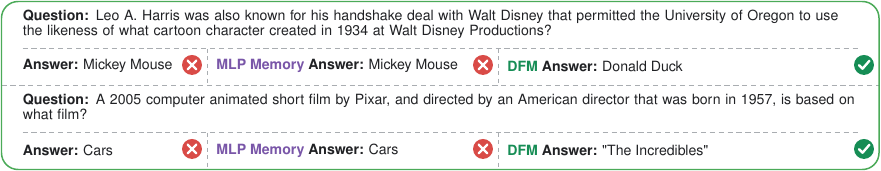}
  \caption{Six selected HotpotQA answer comparisons using Mistral-7B-v0.3. Columns show Base (left), MLP Memory (middle), and DFM (right). Check and cross symbols indicate correct and incorrect normalized exact-match answers.}
  \label{fig:dfm-qa-cases}
\end{figure}

%% file: figures/dfm_retrieval_cases_all7_warm.tex
\begin{figure}[!htbp]
  \centering
  \includegraphics[width=\textwidth]{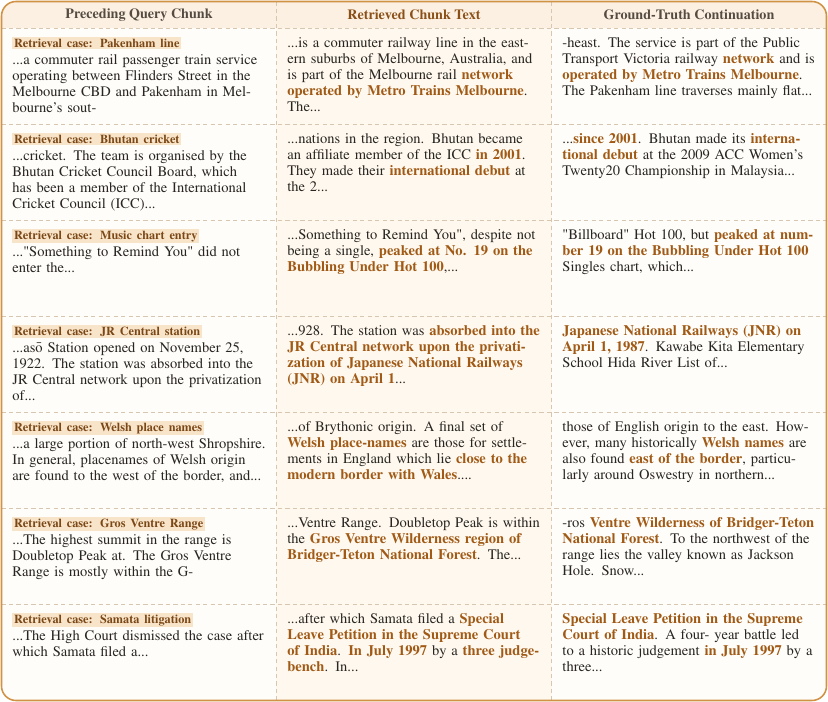}
  \caption{Seven selected offline retrieval examples. Each row shows the preceding query chunk, retrieved chunk text, and ground-truth continuation. Brown spans mark continuation-relevant information in the retrieved text. Text is shortened for display, and split words preserve chunk boundaries.}
  \label{fig:dfm-retrieval-supervision}
\end{figure}